%% file: main.tex
\documentclass[runningheads]{llncs}

\usepackage{booktabs}
\usepackage{multirow}
\usepackage{amsmath}
\usepackage{amssymb}
\usepackage{pifont}
\usepackage{cite}
\usepackage{xcolor}
\usepackage{listings}
\usepackage{booktabs}
\usepackage[table]{xcolor}
\usepackage{colortbl}
\definecolor{headergray}{gray}{0.92}
\definecolor{sectiongray}{gray}{0.96}
\definecolor{bestgreen}{RGB}{0,120,0}
\definecolor{lightblue}{RGB}{235,245,255}
\definecolor{lightgreen}{RGB}{235,250,240}
\definecolor{lightorange}{RGB}{255,245,230}
\definecolor{lightgray}{gray}{0.95}
\definecolor{lightblue}{RGB}{230,242,255}
\newcommand{\cmark}{\textcolor{green!60!black}{\ding{51}}}
\newcommand{\xmark}{\textcolor{red}{\ding{55}}}
\definecolor{headergray}{gray}{0.92}
\definecolor{regionblue}{RGB}{232,244,253}
\definecolor{locgreen}{RGB}{232,250,240}
\definecolor{headergray}{gray}{0.92}
\definecolor{rowblue}{RGB}{235,245,255}
\definecolor{headergray}{gray}{0.92}
\definecolor{rowblue}{RGB}{235,245,255}
\definecolor{bestgreen}{RGB}{0,120,0}
\definecolor{headergray}{gray}{0.92}
\definecolor{sam2blue}{RGB}{235,245,255}
\definecolor{sam3green}{RGB}{235,250,240}
\definecolor{sectiongray}{gray}{0.97}
\definecolor{headergray}{gray}{0.92}
\definecolor{promptone}{RGB}{235,245,255}
\definecolor{prompttwo}{RGB}{235,250,240}
\definecolor{sectiongray}{gray}{0.97}
\definecolor{headergray}{gray}{0.92}
\definecolor{rowblue}{RGB}{235,245,255}
\definecolor{regionblue}{RGB}{232,244,253}
\definecolor{locgreen}{RGB}{232,250,240}
\definecolor{headergray}{gray}{0.92}
\definecolor{rowblue}{RGB}{235,245,255}
\definecolor{regionblue}{RGB}{232,244,253}
\definecolor{locgreen}{RGB}{232,250,240}
\usepackage{hyperref}
\usepackage{caption}
\usepackage{capt-of}
\hypersetup{
    colorlinks=true,
    linkcolor=red,
    citecolor=blue,
    urlcolor=purple
}
\usepackage[T1]{fontenc}

\usepackage{graphicx,verbatim}

\begin{document}
\title{GroundedSurg: A Multi-Procedure Benchmark for Language-Conditioned Surgical Tool Segmentation}

\titlerunning{GroundedSurg}

\author{
Tajamul Ashraf\inst{1}\inst{4}\textsuperscript{\textdagger}\and
Abrar Ul Riyaz*\inst{4} \and
Wasif Tak*\inst{2}\textsuperscript{\textdaggerdbl} \and
Tavaheed Tariq*\inst{4} \and
Sonia Yadav\inst{4} \and
Moloud Abdar\inst{3} \and
Janibul Bashir\inst{4}
}

\renewcommand{\thefootnote}{\textdagger}
\footnotetext{Corresponding author: tajamul.ashraf@kaust.edu.sa}
\renewcommand{\thefootnote}{*}
\footnotetext{Equal Contribution}
\renewcommand{\thefootnote}{\textdaggerdbl}
\footnotetext{Work done while an intern at Gaash Lab, NIT Srinagar}

\authorrunning{Ashraf et al.}

\institute{
King Abdullah University of Science and Technology (KAUST), Saudi Arabia\\
% \email{tajamul.ashraf@kaust.edu.sa}
\and
Thapar Institute of Engineering and Technology, India\\
% \email{wasif.tak@thapar.ac.in}
\and
The University of Queensland, Australia\\
% \email{m.abdar1987@gmail.com}
\and
Gaash Research Lab, National Institute of Technology Srinagar, India\\
% \email{\{abrarnitsri0, wtak732, tawheedtariq090, soniayadavnitsgr, janibbashir\}@nitsri.ac.in}
}
\maketitle              % typeset the header of the contribution
\title{GroundedSurg: A Multi-Procedure Benchmark for Language-Conditioned Surgical Tool Localization}
\vspace{0.35 cm}
\begin{abstract}
Clinically reliable perception of surgical scenes is essential for advancing intelligent, context-aware intraoperative assistance such as instrument handoff guidance, collision avoidance, and workflow-aware robotic support. Existing surgical tool benchmarks primarily evaluate category-level segmentation, requiring models to detect all instances of predefined instrument classes. However, real-world clinical decisions often requires resolving references to a specific instrument instance based on its functional role, spatial relation, or anatomical interaction capabilities not captured by current evaluation paradigms. We introduce \textbf{GroundedSurg}, the first language conditioned, instance-level surgical grounding benchmark. Each instance pairs a surgical image with a natural-language description targeting a single instrument, accompanied by structured spatial grounding annotations including bounding boxes and point-level anchors. The dataset spans ophthalmic, laparoscopic, robotic, and open procedures, encompassing diverse instrument types, imaging conditions, and operative complexities. By jointly evaluating linguistic reference resolution and pixel-level localization, GroundedSurg enables a systematic and realistic evaluation of vision–language models in clinically realistic multi-instrument scenes. Extensive experiments demonstrate substantial performance gaps across modern segmentation and VLMs, highlighting the urgent need for clinically grounded vision–language reasoning in surgical AI systems. Code and data is publically available at \href{https://github.com/gaash-lab/GroundedSurg}{https://github.com/gaash-lab/GroundedSurg}
\end{abstract}
\newpage
\input{introduction}
\input{methdology}

\input{Experiments}
\input{conclusion}

\bibliographystyle{splncs04}
\bibliography{main}

\end{document}

%% file: introduction.tex
\section{Introduction}

Accurate interpretation of surgical scenes has traditionally been framed as the segmentation and recognition of predefined instrument categories~\cite{alabi2025cholecinstanceseg, allan20192017roboticinstrumentsegmentation, maier2017surgical}. These category-level formulations have enabled important downstream applications, including workflow analysis, skill assessment, and context-aware assistance systems~\cite{sachdeva2024phaseinformedtoolsegmentationmanual, LACOSTE, twinanda2016endonetdeeparchitecturerecognition}. However, surgical scene understanding extends beyond class identification. In real operative environments, multiple visually similar instruments
 often coexist within the same field of view. Their clinical relevance depends not only on category, but also on functional role, spatial configuration, and interaction with surrounding anatomy~\cite{inseg, sun2024pixel}. Distinguishing the instrument actively dissecting tissue from another instrument of the same type that is idle or retracting therefore requires resolving a context-dependent reference to a specific instance.

\begin{figure}[t]
    \includegraphics[width=\columnwidth]{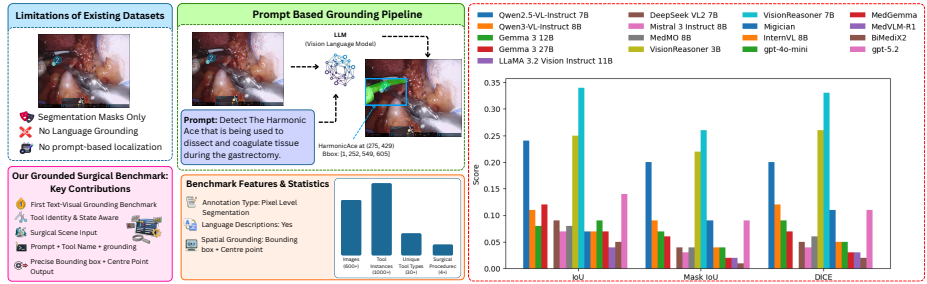}
    \caption{
Overview of \textbf{GroundedSurg}. 
(a) Existing datasets focus on category-level segmentation without language conditioning or instance-level grounding. 
(b) GroundedSurg introduces natural-language queries with structured spatial annotations for query-conditioned instrument localization. 
(c) Baseline results reveal substantial performance gaps, highlighting the challenges of grounded surgical perception.
}
\vspace{-0.4 cm}
\label{figone}
\end{figure}

Current evaluation paradigms do not assess this capability~\cite{inseg, grammatikopoulou2019cadis, sachdeva2024phaseinformedtoolsegmentationmanual}. Category-level tool segmentation benchmarks measure recognition performance but do not require resolving which specific instance satisfies a procedural description when multiple candidates coexist. As summarized in Table~\ref{tab:dataset_stats1}, existing surgical tool datasets primarily focus on class-level segmentation and lack integrated support for language conditioning, structured spatial grounding, and instance-level disambiguation across diverse procedures. Conversely, general vision–language grounding benchmarks (e.g., RefCOCO)~\cite{RefCOCO,RefCOCOg,GRef} do not reflect the visual complexity, occlusions, fine-grained instrument morphology, and domain-specific constraints inherent to surgical environments.
% Conversely, general vision–language benchmarks evaluate referring expression grounding but fail to capture the visual complexity, occlusions, fine-grained instrument morphology, and domain-specific context inherent to surgical environments~\cite{RefCOCO,RefCOCOg,GRef}. 
As a result, grounded, instance-level reasoning in operative scenes remains largely untested. Bridging this gap requires a benchmark that explicitly couples natural-language reference resolution with precise pixel-level instrument localization. Such evaluation must occur under realistic intraoperative conditions, as illustrated in Fig.~\ref{figone}. Such a formulation must support multi-instrument scenes with ambiguity, structured spatial grounding annotations to enforce localization consistency, and evaluation protocols that quantify both grounding accuracy and segmentation precision.

\begin{table}[t]
\caption{
Comparison of surgical tool datasets with respect to grounding capabilities. 
Our benchmark supports multi-procedure, language conditioning, explicit spatial grounding, instance-level disambiguation, and prompt-based evaluation.
}
\label{tab:dataset_stats1}
\centering
\resizebox{\columnwidth}{!}{
\begin{tabular}{lcccccc}
\toprule
\rowcolor{lightgray}
\textbf{Dataset} & \textbf{Domain} & \textbf{Multi-Proc.} & \textbf{Language} & \textbf{Spatial} & \textbf{Disambig.} & \textbf{Prompt} \\
\midrule
CaDIS~\cite{grammatikopoulou2019cadis} & Cataract & \xmark & \xmark & \xmark & \xmark & \xmark \\
AutoLaparo~\cite{wang2022autolaparonewdatasetintegrated} & Laparoscopic & \xmark & \xmark & \xmark & \xmark & \xmark \\
SISVE~\cite{yoon2024sisvseenhanced} & Gastrectomy & \xmark & \xmark & \cmark & \cmark & \xmark \\
EndoVis~\cite{sun2024pixel} & Robotic & \xmark & \xmark & \xmark & \xmark & \xmark \\
CholecSeg8k~\cite{hong2020cholecseg8ksemanticsegmentationdataset} & Cholecystectomy & \xmark & \xmark & \xmark & \xmark & \xmark \\
CholecInstanceSeg~\cite{alabi2025cholecinstanceseg} & Cholecystectomy & \xmark & \xmark & \xmark & \cmark & \xmark \\
Robust-MIS~\cite{ross2020robustmedicalinstrumentsegmentation} & MIS & \xmark & \xmark & \xmark & \xmark & \xmark \\
InSeg1~\cite{inseg} & Ophthalmic & \xmark & \xmark & \cmark & \cmark & \xmark \\
InSeg2~\cite{inseg} & Ophthalmic & \xmark & \xmark & \cmark & \cmark & \xmark \\
\midrule
\rowcolor{lightblue}
\textbf{GroundedSurg (Ours)} & \textbf{Multi-procedure} & \cmark & \cmark & \cmark & \cmark & \cmark \\
\bottomrule
\end{tabular}
}
\vspace{-0.4 cm}
\end{table}

To address this need, we introduce \textbf{GroundedSurg}, a grounding-based surgical tool segmentation benchmark consisting of 612 surgical im-
ages and 1,071 tool-level annotations, that reformulates surgical tool perception as a language-conditioned, instance-level segmentation task. Given a surgical image and a natural-language query describing a specific instrument through its functional role, spatial relation, or anatomical interaction, the objective is to localize and segment the instrument instance satisfying the description. This formulation departs from conventional category-level segmentation by requiring explicit disambiguation among visually similar instruments and grounding of contextual references to a single spatially localized instance.
GroundedSurg incorporates structured spatial annotations, including bounding boxes and center points, to quantify localization accuracy at multiple levels of granularity. Bounding boxes enable coarse instance verification, while pixel-level masks assess fine-grained delineation under realistic surgical challenges such as occlusion, specular reflections, motion blur, and instrument overlap.
 Each image–query pair is treated as an independent evaluation unit, enabling controlled instance-level metric computation in multi-instrument scenes.
By unifying natural-language reference, structured spatial grounding, and multi-procedure diversity within a standardized evaluation framework, GroundedSurg establishes a principled benchmark for grounded surgical perception, thereby enabling the next generation of grounding-aware surgical AI systems.

\noindent\textbf{Contributions.} 
\textbf{(1)} We reconceptualize surgical tool perception as a grounded vision–language task requiring resolution of context-dependent references to specific instrument instances.  
\textbf{(2)} We introduce GroundedSurg, a surgical benchmark that systematically couples natural-language descriptions with explicit spatial grounding annotations, including bounding boxes, center points, and pixel-level masks to enable rigorous evaluation of language-conditioned, instance-level localization and segmentation.
\textbf{(3)} We curate a diverse, multi-procedure dataset spanning heterogeneous surgical domains and imaging conditions, providing a clinically realistic and reproducible testbed for grounding-aware intraoperative vision systems.

%% file: methdology.tex
\section{GroundedSurg}

\textbf{Problem Formulation:}
GroundedSurg formalizes surgical tool perception as a \textit{language-conditioned, instance-level segmentation} task. Each benchmark instance consists of a surgical image paired with a natural-language description referring to a single target instrument, along with structured spatial grounding annotations and a pixel-level segmentation mask.

Formally, let $I \in \mathbb{R}^{H \times W \times 3}$ denote a surgical image of height $H$ and width $W$. For each image, a single target instrument instance is defined by: (i) a natural-language query $T$ describing the instrument through its procedural role, spatial relation, or interaction, (ii) a bounding box $B = (x_{\min}, y_{\min}, x_{\max}, y_{\max})$, (iii) a center point $C = (x_c, y_c)$, and (iv) a binary segmentation mask $M \in \{0,1\}^{H \times W}$. The objective is to predict a segmentation mask $\hat{M}$ corresponding to the instrument described by $T$, i.e., to learn a mapping $f(I, T, B, C) \rightarrow \hat{M}$.
Each image–query pair corresponds to exactly one instrument instance. In scenes containing multiple tools, separate instances are constructed for each reference to ensure unambiguous supervision and controlled instance-level evaluation. The spatial annotations $(B, C)$ act as auxiliary grounding cues to reduce ambiguity and strengthen alignment between linguistic reference and visual localization.
\input{dataset-pipeline}

\noindent\textbf{Benchmark Design:}
Unlike conventional category-level segmentation, GroundedSurg requires models to identify and segment a specific instrument instance conditioned on linguistic and spatial cues, enforcing explicit disambiguation when multiple visually similar tools coexist. The bounding box $B$ provides coarse localization supervision for instance verification, while the center point $C$ introduces an additional spatial constraint to enhance grounding robustness. Despite these anchors, the primary objective remains precise pixel-level delineation of the referenced instrument, reflecting intraoperative accuracy requirements. Treating each image–query pair as an independent evaluation unit prevents cross-instance interference and enables principled instance-level assessment.

\noindent\textbf{Dataset Construction:}
The dataset aggregates samples from publicly available surgical datasets spanning diverse procedures and imaging conditions, including InSeg1~\cite{inseg} and InSeg2~\cite{inseg} (ophthalmic), SISVE~\cite{yoon2024sisvseenhanced} (gastrectomy), EndoVis~\cite{sun2024pixel} (robotic nephrectomy), and CholecInstanceSeg~\cite{alabi2025cholecinstanceseg} (laparoscopic cholecystectomy). These sources cover heterogeneous surgical environments, from microsurgical settings with fine instruments to laparoscopic scenes with articulated tools, cluttered backgrounds, and specular reflections. Such diversity introduces realistic challenges including occlusion, instrument overlap, and high visual similarity, enabling evaluation under clinically representative conditions.

\noindent\textbf{Annotation Protocol:}
Each benchmark instance corresponds to a single tool annotation paired with a grounded natural-language prompt and pixel-level segmentation supervision. When multiple tools appear in the same image, each instance is annotated independently.
\textit{Instance-Level Supervision.}  
Each annotation is assigned a unique identifier. The primary supervision signal is a pixel-level mask precisely delineating the visible extent of the target instrument. All masks are aligned to original image resolution and verified for geometric consistency. Auxiliary spatial grounding information, including bounding box and center point annotations, provides structured localization cues.
\textit{Language Grounding:}  
Queries are generated using Qwen-2.5 VL-Instruct~\cite{qwen2.5} conditioned on visual content and spatial annotations. Each annotation includes (i) a textual description of the target instrument, (ii) explicit spatial grounding via bounding box and center point references, and (iii) a reasoning cue specifying the segmentation objective.

\noindent\textbf{Clinical Validation:}  
The dataset follows a semi-automated, clinically validated pipeline. Initial queries generated by a vision–language model are reviewed and verified by clinicians to eliminate hallucinations, correct semantic inconsistencies, and ensure alignment with surgical context. Refined queries are standardized and integrated into a unified JSON schema, followed by a second round of human validation to guarantee accurate prompt–mask alignment and spatial consistency.

% \textbf{Prompt Types and Reasoning Categories}

% Each prompt refers to a single instrument instance. In multi-tool scenes, separate annotations ensure independent evaluation.
% \textbf{Single-Tool Grounded Segmentation.}  
% Evaluates the ability to associate linguistic descriptions with precise pixel-level segmentation. \textbf{Disambiguation Under Visual Clutter.}  
% Assesses robustness when visually similar distractor instruments are present, requiring resolution via linguistic and spatial cues.

\noindent\textbf{Dataset Statistics:}
\input{dataset_statistics}
The dataset contains approximately \textbf{612 surgical images} and \textbf{1,071 tool-level annotations}. Because multiple instruments may appear per image, annotation count exceeds image count. The dataset spans four procedures and diverse imaging conditions.
All annotations include pixel-level masks, bounding boxes, center points, and natural-language descriptions, establishing a moderate-scale yet fine-grained benchmark for grounding-based surgical segmentation under realistic clinical variability.

\noindent\textbf{Evaluation Protocol:}
Performance is evaluated at the instance level, treating each image–query pair independently.

\textit{Region-Based Metrics: }
Segmentation quality is measured using Intersection over Union (IoU), defined as $\text{IoU} = \frac{|\hat{M} \cap M|}{|\hat{M} \cup M|}$. We report IoU@0.5, IoU@0.9, mean IoU (mIoU), and Dice coefficient to assess overlap and boundary fidelity.

\textit{Localization Metrics: }
Spatial grounding accuracy is evaluated using Bounding Box IoU and Normalized Distance Error (NDE) between predicted and ground-truth center points. All metrics are averaged across instances for systematic comparison of language-conditioned segmentation models in multi-instrument surgical scenes.

%% file: dataset-pipeline.tex
\begin{figure}[t]
    \includegraphics[width=\textwidth]{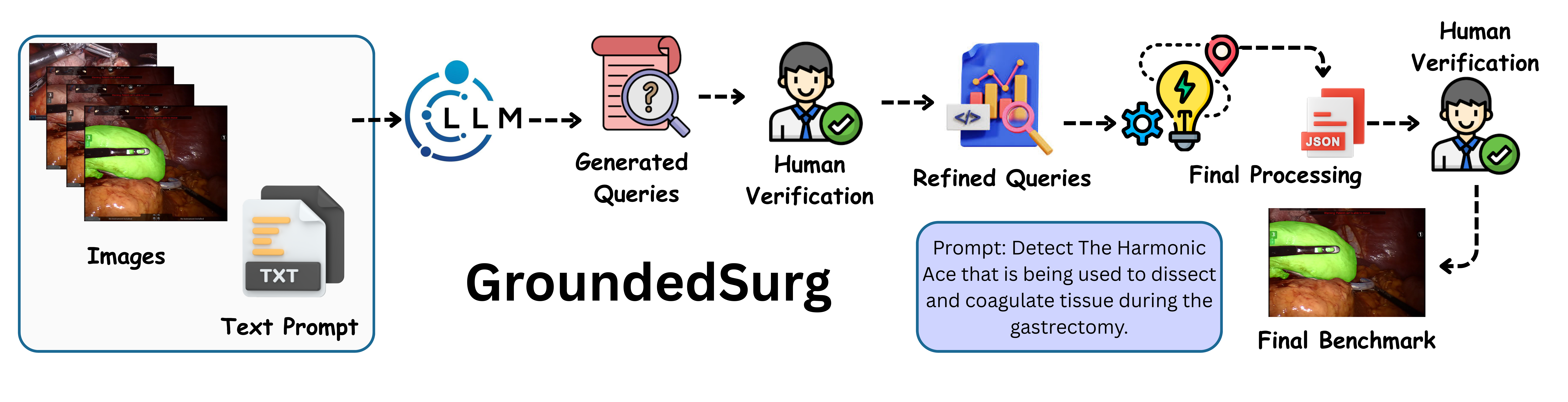}
\caption{
\textbf{GroundedSurg} benchmark pipeline. 
Surgical images are paired with initial prompts and processed by a vision–language model to generate structured instrument descriptions. 
All \textbf{1071} queries are human and clinician verified for semantic correctness and ambiguity removal. 
Final annotations are stored in a standardized JSON schema with spatial grounding (bounding box and center point) and segmentation masks.
}
\vspace{-0.4 cm}
    \label{fig1}
\end{figure}

%% file: dataset_statistics.tex
\begin{table}[t]
\centering
\begin{minipage}[t]{0.47\columnwidth}
\centering
\caption{Summary statistics of the GroundedSurg dataset.}
\label{tab:dataset_stats}
\vspace{2pt}
\resizebox{\linewidth}{!}{
\begin{tabular}{ll}
\toprule
\rowcolor{headergray}
\textbf{Statistic} & \textbf{Value} \\
\midrule
\rowcolor{rowblue}
Number of images & $\sim$612 \\
Number of annotations & $\sim$1{,}071 \\
Average tools/image & $\sim$1.6 \\
Surgical procedures & 4 \\
Annotation type & Pixel-level segmentation \\
Spatial grounding & Box + center point \\
Language descriptions & Instance-level reference \\
\bottomrule
\end{tabular}
}
\end{minipage}
\hspace{0.03\columnwidth}
\begin{minipage}[t]{0.47\columnwidth}
\centering
\caption{Evaluation metrics used in GroundedSurg.}
\label{tab:metrics}
\vspace{2pt}
\resizebox{0.65\linewidth}{!}{
\begin{tabular}{ll}
\toprule
\rowcolor{headergray}
\textbf{Metric} & \textbf{Description} \\
\midrule
\rowcolor{regionblue}
\multicolumn{2}{l}{\textbf{Region-Based Metrics}} \\
IoU & $\frac{|\hat{M} \cap M|}{|\hat{M} \cup M|}$ \\
IoU@0.5/0.9 & IoU $\geq$ 0.5 / 0.9 \\
mIoU & Mean IoU \\
Dice & $\frac{2|\hat{M} \cap M|}{|\hat{M}| + |M|}$ \\
\midrule
\rowcolor{locgreen}
\multicolumn{2}{l}{\textbf{Localization Metrics}} \\
BBox IoU & Box IoU (pred vs GT) \\
NDE & Normalized center distance \\
\bottomrule
\end{tabular}
}
\end{minipage}
\vspace{-0.4 cm}
\end{table}

%% file: Experiments.tex
\section{Experiments}

\noindent\textbf{Experimental Setup:} 
We evaluate GroundedSurg under a unified language-conditioned instance segmentation protocol, where models predict structured localization outputs (bounding box and center point) that are projected onto a frozen SAM-based backend to obtain final masks. 
All models are evaluated in a zero-shot setting without fine-tuning, and performance is reported at the instance level for each image–query pair.

\noindent\noindent\textbf{Comparison Across Vision–Language Models:}
Table~\ref{tab:sam_comparison} presents quantitative comparisons across open-source, reasoning-oriented, medical-domain, and closed-source models under the unified protocol.
\input{comparisons_table}
Across model families, performance remains limited under stricter overlap thresholds. While moderate IoU@0.1 values are observed for certain models, accuracy degrades sharply at higher thresholds (IoU@0.3 and beyond), indicating that coarse localization is occasionally achievable but precise boundary alignment remains challenging.

Among open-source models, Qwen2.5-VL~\cite{qwen2.5} achieves relatively strong coarse grounding performance but exhibits reduced accuracy under stricter overlap criteria, suggesting limited fine-grained spatial precision. In contrast, reasoning-oriented models demonstrate improved localization consistency. VisionReasoner-7B~\cite{liu2025visionreasoner} achieves the highest BBox IoU and Dice scores, indicating stronger spatial grounding and mask fidelity, suggesting that structured reasoning enhances robustness under surgical ambiguity. Medical-domain models do not consistently outperform general-purpose models, indicating that domain pretraining alone does not guarantee improved instance-level grounding. Closed-source systems achieve competitive but not dominant performance, reinforcing the overall difficulty of the benchmark.

\noindent\textbf{Segmentation Backend Analysis.}
To assess the impact of the promptable segmentation backend, we compare representative models under different frozen segmentation models.
\input{sam2_vs_sam3}
Table~\ref{tab:sam_comparison} shows that segmentation quality varies substantially between SAM2~\cite{ravi2024sam2segmentimages} and SAM3~\cite{jiang2026medicalsam3}. While some models benefit from improved mask projection under SAM3, others show marginal gains. Notably, VisionReasoner exhibits a pronounced improvement under SAM3, suggesting stronger compatibility between accurate localization outputs and advanced mask decoding. These results highlight the tight coupling between grounding accuracy and segmentation projection in the overall pipeline.

\input{change_prompt}

As shown in Table~\ref{tab:change_prompt}, general-purpose vision–language models exhibit significant performance variation across prompt styles. For example, Qwen3-VL~\cite{qwen3} and several other models experience notable drops in IoU and BBox IoU under the alternate prompt. In contrast, reasoning-oriented models demonstrate greater robustness; VisionReasoner-7B maintains and even improves performance under Prompt2, indicating stronger semantic grounding and reduced reliance on rigid prompt structure. These findings suggest that prompt engineering remains critical for general-purpose multimodal models, whereas reasoning-focused architectures exhibit improved invariance to linguistic rephrasing.

\noindent\textbf{Qualitative Analysis.}
\input{ablations}
Figure~\ref{fig1} presents qualitative comparisons across representative models. General-purpose vision–language models, which do not directly perform segmentation, often generate coarse or inaccurate spatial localizations in cluttered surgical scenes. When these predicted regions are subsequently provided to SAM3 for mask generation, the resulting segmentations remain imprecise, with inaccurate boundaries and contextual leakage—particularly in multi-instrument settings. These qualitative observations are consistent with the quantitative trends reported in Table~\ref{tab:sam_comparison}.

% Figure~\ref{fig1} presents qualitative comparisons across representative models. General-purpose models  frequently over-segment background regions or mislocalize instruments in cluttered scenes, whereas reasoning-oriented models produce more spatially precise masks with reduced contextual leakage, particularly in multi-instrument settings. These visual examples align with the quantitative trends in Table~\ref{tab:sam_comparison}.

\noindent\textbf{Effect of Prompt Tuning.}
The prompt sensitivity results (Table~\ref{tab:change_prompt}) confirm that segmentation performance is strongly influenced by instruction structure. While many models show instability under minor rephrasing, reasoning-oriented models maintain consistent grounding behavior, indicating that explicit spatial reasoning improves robustness to linguistic variability.

\noindent\textbf{Segmentation Projection Effects.}
The comparison between SAM2 and SAM3 (Table~\ref{tab:sam_comparison}) further shows that mask quality depends not only on localization accuracy but also on backend projection characteristics. Improvements in mask decoding disproportionately benefit models with stronger localization, emphasizing the importance of evaluating grounding and segmentation components.

%% file: comparisons_table.tex
\begin{table}[t]
\centering
\caption{
Quantitative comparison of vision-language models on GroundedSurg under a unified query-conditioned segmentation protocol.
Best results are shown in \textbf{bold}, second-best are \underline{underlined}.
}
\label{tab:sam_comparison}
\resizebox{\columnwidth}{!}{
\begin{tabular}{lccccccccc}
\toprule
\rowcolor{headergray}
Model & Params & IoU@0.1 & IoU@0.3 & BBox IoU & mIoU@0.1 & mIoU@0.3 & Mask IoU & NDE $\downarrow$ & Dice \\
\midrule

\rowcolor{sectiongray}
\multicolumn{10}{l}{\textit{Open-Source Models}} \\

Qwen2.5-VL-Instruct~\cite{qwen2.5} & 7B & \textbf{0.52} & \textbf{0.33} & \underline{0.24} & \textbf{0.26} & \textbf{0.22} & \underline{0.20} & \underline{1.45} & \underline{0.20} \\
Qwen3-VL-Instruct~\cite{qwen3} & 8B & 0.36 & 0.13 & 0.11 & 0.17 & 0.12 & 0.09 & 1.78 & 0.12 \\
Gemma 3~]\cite{gemmateam2025gemma3technicalreport} (12B) & 12B & 0.28 & 0.09 & 0.08 & 0.13 & 0.09 & 0.07 & \underline{1.45} & 0.09 \\
Gemma 3~\cite{gemmateam2025gemma3technicalreport} (27B) & 27B & \underline{0.43} & 0.11 & 0.12 & 0.11 & 0.07 & 0.06 & 1.52 & 0.07 \\
LLaMA 3.2 Vision~\cite{grattafiori2024llama} & 11B & 0.02 & 0.00 & 0.00 & 0.01 & 0.00 & 0.00 & 2.90 & 0.00 \\
DeepSeek VL2~\cite{deepseek_vl2} & 7B & 0.28 & 0.10 & 0.09 & 0.06 & 0.05 & 0.04 & 1.89 & 0.05 \\
Mistral 3~\cite{liu2026ministral} & 8B & 0.27 & 0.06 & 0.07 & 0.07 & 0.04 & 0.03 & 1.89 & 0.04 \\

\midrule
\rowcolor{sectiongray}
\multicolumn{10}{l}{\textit{Reasoning-Oriented Models}} \\

VisionReasoner (3B)~\cite{liu2025visionreasoner} & 3B & 0.20 & 0.06 & 0.25 & 0.23 & 0.03 & 0.22 & 1.39 & 0.26 \\
VisionReasoner (7B)~\cite{liu2025visionreasoner} & 7B & 0.32 & \underline{0.20} & \textbf{0.34} & \underline{0.25} & \underline{0.12} & \textbf{0.26} & \textbf{1.05} & \textbf{0.33} \\
Migician~\cite{li2025migician} & 7B & 0.19 & 0.08 & 0.07 & 0.16 & 0.12 & 0.09 & 1.72 & 0.11 \\
InternVL~\cite{chen2024internvl} & 8B & 0.29 & 0.05 & 0.07 & 0.08 & 0.05 & 0.04 & 1.69 & 0.05 \\

\midrule
\rowcolor{sectiongray}
\multicolumn{10}{l}{\textit{Medical-Domain Models}} \\

MedMO~\cite{deria2026medmo} & 8B & 0.33 & 0.04 & 0.08 & 0.08 & 0.05 & 0.04 & 1.50 & 0.06 \\
MedGemma~\cite{sellergren2025medgemma} & 4B & 0.29 & 0.04 & 0.07 & 0.05 & 0.03 & 0.02 & 1.88 & 0.03 \\
MedVLM-R1~\cite{pan2025medvlm} & 2B & 0.13 & 0.01 & 0.04 & 0.05 & 0.03 & 0.02 & 2.16 & 0.03 \\
BiMediX2~\cite{mullappilly2024bimedix2biomedicalexpertlmm} & 8B & 0.23 & 0.04 & 0.05 & 0.07 & 0.02 & 0.01 & 2.11 & 0.02 \\

\midrule
\rowcolor{sectiongray}
\multicolumn{10}{l}{\textit{Closed-Source Models}} \\

GPT-4o-mini~\cite{gpt4} & -- & 0.35 & 0.04 & 0.09 & 0.08 & 0.05 & 0.04 & 1.63 & 0.05 \\
GPT-5.2~\cite{singh2025openai} & -- & 0.39 & \underline{0.20} & 0.14 & 0.16 & 0.12 & 0.09 & 1.95 & 0.11 \\

\bottomrule
\end{tabular}
}
\vspace{-0.5 cm}
\end{table}

%% file: sam2_vs_sam3.tex
\begin{figure*}[t]
\centering

% -------- LEFT (TABLE) --------
\begin{minipage}[t]{0.47\textwidth}
\centering
\small
\vspace{-4 cm}
\textbf{Table 5.} Comparison of representative models under SAM2~\cite{ravi2024sam2segmentimages} and SAM3~\cite{jiang2026medicalsam3} as segmentation models after detections from VLMS. 
Best results are highlighted in \textbf{bold}, second-best are \underline{underlined}.

\vspace{4pt}

\resizebox{\linewidth}{!}{
\begin{tabular}{lcccc}
\toprule
Model & \multicolumn{2}{c}{SAM2} & \multicolumn{2}{c}{SAM3} \\
\cmidrule(lr){2-3}\cmidrule(lr){4-5}
& Mask IoU & Dice & Mask IoU & Dice \\
\midrule
Qwen2.5 & \textbf{0.22} & \textbf{0.26} & \underline{0.20} & 0.20 \\
Qwen3 & 0.12 & 0.16 & 0.09 & 0.12 \\
Gemma 3 27B & 0.03 & 0.04 & 0.06 & 0.07 \\
MedMO & 0.00 & 0.01 & 0.04 & 0.06 \\
gpt-5.2 & 0.04 & 0.06 & 0.09 & 0.11 \\
VR-7B & \underline{0.17} & \underline{0.19} & \textbf{0.26} & \textbf{0.33} \\
\bottomrule
\end{tabular}
}

\end{minipage}
\hfill
% -------- RIGHT (FIGURE) --------
\begin{minipage}[t]{0.52\textwidth}
\centering
\includegraphics[width=\linewidth]{components.pdf}
\label{gsurg_components}
\vspace{4pt}
\small
\textbf{Fig. 3.} GroundedSurg components.
\end{minipage}
\vspace{-0.5 cm}
\end{figure*}

%% file: change_prompt.tex
\begin{table}[t]
\centering
\caption{
Sensitivity analysis under two prompt formulations.
Best results per metric are shown in \textbf{bold}, second-best are \underline{underlined}.
}
\label{tab:change_prompt}

\resizebox{\columnwidth}{!}{
\begin{tabular}{lcccccccc}
\toprule
\rowcolor{headergray}
\multirow{2}{*}{\textbf{Model}} 
& \multicolumn{4}{c}{\textbf{Prompt1}} 
& \multicolumn{4}{c}{\textbf{Prompt2}} \\
\cmidrule(lr){2-5} \cmidrule(lr){6-9}

& \textbf{IoU@0.1} & \textbf{IoU@0.3} & \textbf{BBox IoU} & \textbf{NDE $\downarrow$}
& \textbf{IoU@0.1} & \textbf{IoU@0.3} & \textbf{BBox IoU} & \textbf{NDE $\downarrow$} \\
\midrule

\rowcolor{promptone}
Qwen2.5-VL-7B~\cite{qwen2.5} 
& \underline{0.52} & \underline{0.33} & 0.24 & 1.45 
& 0.38 & 0.09 & 0.11 & \underline{1.43} \\

Qwen3-VL-8B~\cite{qwen3}
& 0.36 & 0.13 & 0.11 & 1.78 
& 0.11 & 0.02 & 0.09 & 2.94 \\

Gemma3-27B~\cite{gemmateam2025gemma3technicalreport} 
& 0.43 & 0.11 & 0.12 & 1.52 
& 0.42 & 0.08 & 0.10 & 1.51 \\

InternVL-8B~\cite{chen2024internvl} 
& 0.29 & 0.05 & 0.07 & 1.69 
& 0.37 & 0.09 & 0.10 & 1.42 \\

MedGemma-4B-it~\cite{sellergren2025medgemma} 
& 0.29 & 0.04 & 0.07 & 1.88 
& 0.21 & 0.06 & 0.07 & 1.91 \\

\rowcolor{prompttwo}
VisionReasoner-7B~\cite{liu2025visionreasoner} 
& 0.32 & 0.08 & \textbf{0.34} & \textbf{1.05} 
& \textbf{0.58} & \textbf{0.39} & \underline{0.31} & 1.14 \\

\bottomrule
\end{tabular}
}
\vspace{-0.5 cm}
\end{table}

%% file: ablations.tex
\begin{figure}[t]
    \includegraphics[width=\textwidth]{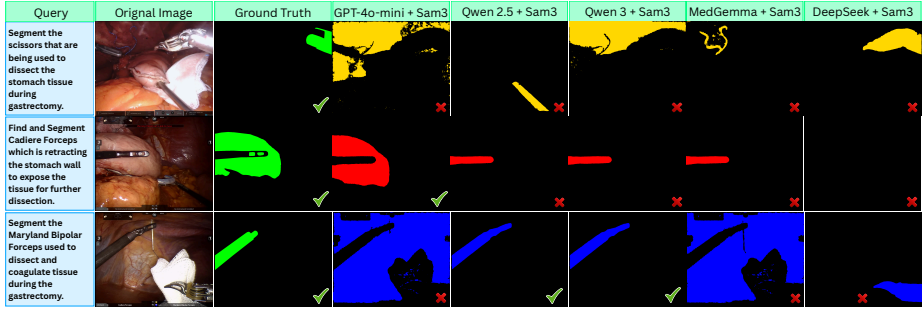}
\caption{Qualitative comparison on GroundedSurg showing that reasoning-oriented models produce more spatially precise masks than general-purpose models when projecting structured localization outputs onto a frozen segmentation backend, particularly in multi-instrument and visually cluttered scenes.}
    \label{fig1}
    \vspace{-0.5 cm}
\end{figure}

%% file: conclusion.tex
\section{Conclusion}

We introduced GroundedSurg, a grounding-based benchmark that reformulates surgical instrument perception as a language-conditioned, instance-level segmentation task. By integrating natural-language references with structured spatial grounding and pixel-level masks, we advance beyond category-level recognition toward context-dependent, clinically meaningful localization across diverse procedures.
Experiments show that current multimodal models struggle with reliable instance-level grounding in complex surgical scenes. While coarse localization is occasionally achievable, precise boundary delineation and robustness to prompt variation remain limited. 
GroundedSurg provides a standardized and clinically relevant testbed for advancing grounding-aware intraoperative vision and highlights the need for models that better integrate linguistic reasoning with fine-grained spatial perception.